\documentclass{article}
\usepackage{tabularx}

\usepackage[dblblindworkshop, final]{neurips_2025}
\usepackage{wrapfig}
\usepackage{booktabs}  

\usepackage[utf8]{inputenc} 
\usepackage[T1]{fontenc}    
\usepackage{hyperref}       
\usepackage{url}            
\usepackage{booktabs}       
\usepackage{amsfonts}       
\usepackage{nicefrac}       
\usepackage{microtype}      
\usepackage{xcolor}         

\usepackage{subcaption} 
\usepackage{tabularx}

\usepackage{bbm}
\usepackage{amsmath} 
\usepackage{multirow} 
\usepackage{enumitem}

\usepackage{algorithm}
\usepackage{algpseudocode}
\algtext*{EndIf}
\algtext*{EndFor}
\algtext*{EndWhile}

\usepackage{listings}
\usepackage{amsthm}

\usepackage{xcolor}
\lstdefinestyle{kgprompt}{
  backgroundcolor=\color{gray!10},    
  frame=single,                       
  rulecolor=\color{gray!70},          
  basicstyle=\ttfamily\small,         
  breaklines=true,                    
  columns=fullflexible,
  keepspaces=true,
  showstringspaces=false,
}
\usepackage{times}
\usepackage{latexsym}
\usepackage{booktabs}

\usepackage{url}
\usepackage{todonotes}
\usepackage{amssymb}
\usepackage{amsmath}
\usepackage{algorithm}
\usepackage{floatflt}   
\usepackage{graphicx}   

\usepackage[T1]{fontenc}

\usepackage[utf8]{inputenc}

\usepackage{microtype}

\usepackage{inconsolata}
\usepackage{tikz}
\usetikzlibrary{positioning}  

\workshoptitle{Scaling Environments for Agents (SEA)}

\title{GR-Agent: Adaptive Graph Reasoning Agent under Incomplete Knowledge}
\author{%
Dongzhuoran Zhou\textsuperscript{1,2},
Yuqicheng Zhu\textsuperscript{2,3}, 
Xiaxia Wang\textsuperscript{4},
Hongkuan Zhou\textsuperscript{2,3}, \\
\textbf{Jiaoyan Chen\textsuperscript{6}},
\textbf{Steffen Staab\textsuperscript{3,7}},
\textbf{Yuan He\textsuperscript{5,4}\thanks{Shared supervision.}},
\textbf{Evgeny Kharlamov\textsuperscript{1,2}\footnotemark[1]} 
\\
\textsuperscript{1}University of Oslo, 
\textsuperscript{2}Bosch Center for AI, 
\textsuperscript{3}University of Stuttgart, 
\textsuperscript{4}University of Oxford, \\
\textsuperscript{5}Amazon, 
\textsuperscript{6}The University of Manchester, 
\textsuperscript{7}University of Southampton\\
\texttt{dongzhuoran.zhou@de.bosch.com}\\
}
\begin{document}

\maketitle

\begin{abstract}
Large language models (LLMs) achieve strong results on knowledge graph question answering (KGQA), but most benchmarks assume complete knowledge graphs (KGs) where direct supporting triples exist. This reduces evaluation to shallow retrieval and overlooks the reality of incomplete KGs, where many facts are missing and answers must be inferred from existing facts.  
We bridge this gap by proposing a methodology for constructing benchmarks under KG incompleteness, which removes direct supporting triples while ensuring that alternative reasoning paths required to infer the answer remain. Experiments on benchmarks constructed using our methodology show that existing methods suffer consistent performance degradation under incompleteness, highlighting their limited reasoning ability.  
To overcome this limitation, we present the Adaptive Graph Reasoning Agent (GR-Agent). It first constructs an interactive environment from the KG, and then formalizes KGQA as agent environment interaction within this environment. 
GR-Agent operates over an action space comprising graph reasoning tools and maintains a memory of potential supporting reasoning evidence, including relevant relations and reasoning paths. Extensive experiments demonstrate that GR-Agent outperforms non-training baselines and performs comparably to training-based methods under both complete and incomplete settings.
\end{abstract}


\section{Introduction}

Large language models (LLMs) have demonstrated impressive capability across a wide range of knowledge-intensive tasks \citep{lewis2020retrieval, Khandelwal2020rag, Izacard2021FiD, Borgeaud2022retro, ram2023ICRALM, zhou2025gqc}.
One representative task is knowledge graph question answering (KGQA), which aims to answer natural language questions using the information provided by a knowledge graph (KG)~\citep{Yih2016webqsp,gu2021beyond,ye2021rng,yao2019kg,baek2023direct}. A growing body of work suggests that LLMs can be effective at retrieving relevant triples and reasoning over them to answer questions \citep{linhao2024rog, he2024g, mavromatis2024gnn, chen2024pog, Jiashuo2024ToG, jiang2023structgpt}.

However, existing benchmarks fall short of properly testing this reasoning ability.
Specifically, current datasets are constructed under the assumption that the KG is complete, i.e., every question has direct supporting triples in the KG \citep{Yih2016webqsp, talmor2018web, gu2021beyond}. This design reduces the task to a retrieval problem: \emph{models can answer correctly simply by locating the explicit facts in the KG}.
For instance, consider the question ``Who is Justin Bieber’s uncle?''. In a complete KG, this might be directly answered by a triple such as $\langle\texttt{Justin Bieber}, \texttt{hasUncle}, \texttt{Brad Bieber}\rangle$. 
Such evaluations therefore conflate true reasoning with shallow retrieval and fail to reflect the reality of incomplete KGs, where \emph{many facts are missing and questions must be answered by reasoning over existing facts}.
In this case, if $\langle\texttt{Justin Bieber}, \texttt{hasUncle}, \texttt{Brad Bieber}\rangle$ is missing, the answer must be inferred by combining $\langle\texttt{Brad Bieber}, \texttt{hasSon}, \texttt{Jaxon Bieber}\rangle$ with $\langle\texttt{Jaxon Bieber}, \texttt{hasBrother}, \texttt{Justin Bieber}\rangle$.

To enable systematic evaluation under this realistic setting, we first propose a general method for constructing benchmarks datasets in which each question requires genuine reasoning.
That is, for every question, the direct supporting triples are removed from the KG, ensuring that the answer can only be inferred via high-confidence logical rules rather than shallow retrieval.
We then evaluate six representative baseline methods on the resulting benchmark, and observe consistent performance degradation across all models when moving from the complete to the incomplete setting. This demonstrates that current approaches predominantly rely on shallow retrieval rather than reasoning.


To address this challenge, we propose the Adaptive Graph Reasoning Agent (GR-Agent), 
a training-free agentic framework specifically designed to reason over KGs. 
We construct an interactive environment on top of the KG, which consists of:
(i) a mutable state represented by a memory that stores explored paths, grounded reasoning paths, and entities; 
(ii) an action space comprising reasoning tools for relation-path exploration, reasoning-path grounding, and answer synthesis;
and (iii) transition and observation mechanisms that update the 
memory and expose new candidates during interaction. 
Through this interaction, the agent adaptively expands the search frontier, prioritizes promising paths, and synthesizes a final answer.
A comprehensive empirical study substantiates our claims, showing that 
GR-Agent consistently outperforms non-training baselines and achieves performance comparable 
to training-based baselines across both complete and incomplete settings.



Our contributions are three-fold: \textbf{First}, we propose a general methodology for constructing benchmarks under incompleteness and instantiate it on the Family and FB15k-237 KGs, providing systematic testbeds for evaluating reasoning ability of LLMs.  \textbf{Second}, we highlight the challenge posed by KG incompleteness and empirically demonstrate its impact on the reasoning ability of existing methods. \textbf{Third}, we introduce GR-Agent, a novel training-free agentic framework for KGQA task. 

\section{Related Works}

\paragraph{KGQA}
The KGQA task aims to answer natural language questions using the KG~$\mathcal{G}$, where $\mathcal{G}$ 
is represented as a set of binary facts $r(s, o)$, with $r\in\mathcal{R}$ denoting a predicate and $s, o\in\mathcal{E}$ denoting entities. 
The answer to each question is one or more entities in $\mathcal{G}$. 
Approaches to KGQA fall into three categories: 
(1) \textit{Semantic parsing-based methods} \citep{Yih2016webqsp,gu2021beyond,ye2021rng}, which translate 
questions into formal queries (e.g., SPARQL) and execute them over the KG. They provide high precision 
and interpretability but struggle with complex language,  diverse logical forms, large parsing 
search spaces \citep{lan2021survey} and mismatch between query and KG construction \citep{zhou2022ontology}. 
(2) \textit{Embedding-based methods} \citep{yao2019kg,baek2023direct}, which encode questions and entities 
into a shared space and rank candidates by similarity. They are end-to-end and annotation-free but face 
challenges in multi-hop reasoning \citep{qiu2020stepwise,zhou2025glora}, interpretability~\citep{biswas2023knowledge}, 
and prediction uncertainty~\citep{zhu-etal-2024-predictive,zhu2025conformalized,zhu-etal-2025-predicate, zhu2025unkgcp}. 
(3) \textit{RAG-based methods}, which go beyond previous approaches by coupling retrieval with 
the generative reasoning capabilities of LLMs \citep{linhao2024rog, he2024g, mavromatis2024gnn, chen2024pog}.

\paragraph{KGQA Benchmarks}
Benchmarks like WebQSP~\citep{Yih2016webqsp}, CWQ~\citep{talmor2018web}, and GrailQA~\citep{gu2021beyond} assume complete KGs by construction, retaining only questions answerable via direct supporting triples. To study incompleteness, recent work deletes triples randomly or along shortest paths~\cite{xu2024generate,zhou2025evaluating}, but this often leaves questions unanswerable, conflating missing knowledge with model reasoning limits.

\paragraph{KGQA with Agents}
Building on RAG-based approaches, a new line of work adopts an \emph{agentic perspective}, where the 
LLM is treated as an autonomous agent that iteratively interacts with the KG. 
Representative examples include ToG~\citep{Jiashuo2024ToG}, StructGPT~\citep{jiang2023structgpt}, 
KG-Agent~\citep{jiang2024kg}, and GoG~\citep{xu2024generate}. 
While these approaches highlight the promise of LLMs as reasoning agents, they often rely on fixed 
exploration ranges, assume complete KGs, or defer reasoning to the LLM’s internal knowledge. 
In contrast, our work introduces an agent that formalizes the KG as an environment and reasons 
directly at the relation-path level, enabling multi-hop reasoning over both complete and incomplete KGs.


\section{Benchmark Construction}
\label{sec:da_construct}
This section introduces a general method for constructing benchmarks to evaluate KGQA under varying degrees of knowledge incompleteness \citep{zhou2025breaks}.
The key objective is to create natural language questions whose answers are not directly stated in the KG but can be logically inferred through reasoning over alternative paths.

To achieve this, we first mine high-confidence logical rules from the KG to identify triples that are inferable via reasoning. We then remove a subset of these triples while preserving the supporting facts required for inference. Natural language questions are generated based on the removed triples, meaning that models must rely on reasoning rather than direct retrieval to answer the questions.

\subsection{Rule Mining}
\label{sec:rule_mining}
To ensure that questions in our benchmark require reasoning rather than direct lookup, we first identify triples that are logically inferable from other facts.
We achieve this by mining high-confidence Horn rules~\citep{horn1951sentences} from the original KG using the \textsc{AMIE3} algorithm~\citep{lajus2020fast}.

\textsc{AMIE3} is a widely used rule mining system designed to operate efficiently over large-scale KGs. 
A logical rule discovered by \textsc{AMIE3} has the following form:
\begin{equation*}
B_1 \wedge B_2 \wedge \dots \wedge B_n \Rightarrow H\,,
\label{eq:horn}
\end{equation*}
where $B_i$ are body atoms and $H$ is the head atom. For example:
\begin{equation*}
\texttt{hasParent}(X,Y)\wedge \texttt{hasSibling}(Y,Z)\Rightarrow \texttt{hasUncle}(X,Z)\,,
\label{eq:horn}
\end{equation*}
expressing that if $Y$ is a parent of $X$ and $Y$ has a sibling $Z$, then $Z$ is likely an uncle of $X$.

A \emph{grounding} of this rule substitutes entities in the KG for variables (e.g., $X{=}$\texttt{Justin}, $Y{=}$\texttt{Mary}, $Z{=}$\texttt{John}). A rule is considered \emph{well supported} if many such groundings exist in the KG, and \emph{confidence} is measured by the proportion for all grounded rule bodies in the KG, their grounded head atom also exists in the KG. 
Only high-confidence and well-supported rules are retained. Detailed metrics and filtering criteria are provided in Appendix~\ref{app:rule_mining}.

\subsection{Dataset Generation}
\label{sec:dataset_generation}
We aim to generate questions that cannot be answered using direct supporting triples, but for which sufficient information is implicitly available in the KG. The core idea is to first remove triples that can be reliably inferred using high-confidence rules mined by \textsc{AMIE3}, and then generate questions based on these removed triples.

\begin{itemize}
    \item \textbf{Triple Removal.} For each mined rule, we select up to 30 groundings where both body and head triples exist. The head triple is then removed while preserving all body triples, ensuring that each removed fact remains inferable and that no body triple required for other groundings is lost.
    \item \textbf{Question Generation.} For each removed triple, we prompt GPT-4 to generate a natural-language question asking for the answer entity based on the predicate and a specified topic entity. To encourage diversity, either the head or tail entity is randomly chosen as the topic, with the other as the answer. The prompt template is given in Appendix~\ref{app:generation_prompt}.
    \item \textbf{Answer Set Completion.} Although each question is initially generated based on a single deleted triple, there may exist multiple correct answers in the KG. To ensure rigorous and unbiased evaluation, we construct for each question a complete set of correct answers and mark the answers requiring inference (i.e., without direct supporting triples) as "hard". 
\end{itemize}





\subsection{Dataset Overview}
\label{sec:data_overview}
\paragraph{KGs.} To support a systematic evaluation of reasoning under knowledge incompleteness, we construct benchmark datasets based on two well established KGs: \textbf{Family} \citep{sadeghian2019drum} and \textbf{FB15k-237} \citep{toutanova2015observed}. These datasets differ in size, structure, and domain coverage, enabling evaluation across both synthetic and real-world settings.

\paragraph{Mined Rules.}
Table~\ref{tab:rule-stats} summarizes the number of mined rules for each dataset, categorized by rule type.
The listed types (e.g., symmetry, inversion, composition) correspond to common logical patterns,
while the \emph{other} category includes more complex or irregular patterns
(See Appendix~\ref{app:other_rules} for details).

\begin{table*}[t]
  \centering
  \begin{minipage}[b]{0.48\textwidth}
    \centering
    \resizebox{\linewidth}{!}{%
    \begin{tabular}{lrr}
      \toprule
      \textbf{Rule Type} & \textbf{Family} & \textbf{FB15k-237} \\
      \midrule
      Symmetry: $r(x,y)\Rightarrow r(y,x)$         & 0    & 27   \\
      Inversion: $r_1(x,y)\Rightarrow r_2(y,x)$    & 6    & 50   \\
      Hierarchy: $r_1(x,y)\Rightarrow r_2(x,y)$    & 0    & 76   \\
      Composition: $r_1(x,y)\wedge r_2(y,z)\Rightarrow r_3(x,z)$ & 56 & 343 \\
      Other                                        & 83   & 570  \\
      \midrule
      \textbf{Total} & \textbf{145} & \textbf{1,066} \\
      \bottomrule
    \end{tabular}}
    \caption{Statistics of mined rules.}
    \label{tab:rule-stats}
  \end{minipage}\hfill
  \begin{minipage}[b]{0.5\textwidth}
    \centering
    \renewcommand{\arraystretch}{1.3}  
    \resizebox{\linewidth}{!}{%
    \begin{tabular}{lccccc}
      \toprule
      Dataset & \#Triples & Train & Val & Test & Total Qs \\
      \midrule
      Family-Complete      & 17,615  & 1,749 & 218 & 198 & 2,165 \\
      Family-Incomplete    & 15,785  & 1,749 & 218 & 198 & 2,165 \\
      FB15k-237-Complete   & 204,087 & 4,374 & 535 & 540 & 5,449 \\
      FB15k-237-Incomplete &198,183  & 4,374 & 535 & 540 & 5,449 \\
      \bottomrule
    \end{tabular}}
    \caption{Dataset statistics.}
    \label{tab:data-stats}
  \end{minipage}
\end{table*}

\begin{table*}[h]
\centering

\resizebox{\linewidth}{!}{
\begin{tabular}{@{}ll@{}}
\toprule
\textbf{Dataset} & \textbf{Example} \\ 
\midrule

\textbf{Family} & 
\begin{tabular}[t]{@{}l@{}}
\textit{Question:} Who is 139's brother? \quad \textit{Topic Entity:} 139 \\
---\\
\textit{Answer:} [\textcolor{red}{205}, 138, 2973, 2974] \\
\textit{Direct Supporting Triples:} \texttt{brotherOf(139,205)}  \\
\textit{Alternative Paths:} \texttt{fatherOf(139,14) $\wedge$ uncleOf(205,14) $\Rightarrow$ brotherOf(139,205)}  \\
\end{tabular} \\

\midrule

\textbf{FB15k-237} & 
\begin{tabular}[t]{@{}l@{}}
\textit{Question:} What is the currency of the estimated budget for 5297 (Annie Hall)? \quad \textit{Topic Entity:} 5297 (Annie Hall) \\
--- \\
\textit{Answer:} [\textcolor{red}{1109 (United States Dollar)}]  \\
\textit{Direct Supporting Triples:} \texttt{filmEstimatedBudgetCurrency(5297, 1109)}  \\
\begin{tabular}{@{}r@{\;}l@{}}
\textit{Alternative Paths:} & \texttt{filmCountry(5297 (Annie Hall), 2896 (United States of America))} \\
$\wedge$ & \texttt{locationContains(2896 (United States of America), 9397 (New York))} \\
$\wedge$ & \texttt{statisticalRegionGdpNominalCurrency(9397 (New York), 1109 (United States Dollar))} \\
$\Rightarrow$ & \texttt{filmEstimatedBudgetCurrency(5297 (Annie Hall), 1109 (United States Dollar))}
\end{tabular}
\end{tabular} \\

\bottomrule
\end{tabular}
}
\caption{Examples from our benchmark datasets. 
Each instance includes a \emph{natural-language question}, a \emph{topic entity}, and the full \emph{set of correct answers}. The 
\textcolor{red}{red-highlighted} 
answer denotes the \emph{hard answer}, i.e., the one whose supporting triple has been removed in the incomplete KG setting. We also show the corresponding \emph{direct supporting triples} (the deleted triple) and an \emph{alternative reasoning path} derived from a mined rule that enables inference of the answer.
}
\label{tab:qa-examples}
    \vspace{-5mm} 

\end{table*}

\paragraph{Datasets.}
Each dataset instance consists of (1) a natural-language question,
(2) a topic entity referenced in the question,
and (3) a complete set of correct answer entities derived from the original KG. Table~\ref{tab:qa-examples} presents representative examples from each dataset.
The final question set is randomly partitioned into training, validation, and test sets using an 8:1:1 ratio.
This split is applied uniformly across both datasets to ensure consistency.

We provide two retrieval sources per dataset: 
(1) \textbf{Complete KG}: the original KG containing all triples.
(2) \textbf{Incomplete KG}: a modified version where selected triples, deemed logically inferable via \textsc{AMIE3}-mined rules, are removed (Section~\ref{sec:dataset_generation}).

Table~\ref{tab:data-stats} summarizes the number of KG triples and generated questions in each split for both datasets, under complete and incomplete KG settings.

\subsection{Evaluation Protocol}

\paragraph{Evaluation Setup.}
Given a natural language question $q\in\mathcal{Q}$, access to a KG $\mathcal{G}$, and a topic entity, the model is designed to return a set of predicted answer entities $\mathcal{P}_q$. 
Since LLMs typically produce raw text sequences as output, we extract the final prediction set $\mathcal{P}_q$ by applying string partitioning and normalizing, following \citet{linhao2024rog}. 
Details of this postprocessing step are provided in Appendix \ref{app:eval_setting}.
Without specific justification, all entities are represented by randomly assigned indices without textual labels (e.g., “\texttt{Barack Obama}” becomes \texttt{39}) to ensure that models rely solely on knowledge from the KG rather than memorized surface forms.

\paragraph{Evaluation Metrics.}
Given a set of questions $\mathcal{Q}$, 
we denote the predicted answer set $\mathcal{P}_q$ and the gold answer set $\mathcal{A}_q$, respectively, for each question $q\in\mathcal{Q}$.
The evaluation metrics are defined as follows:
\begin{itemize}
    \item \textbf{Hits@\!Any.}  
        Hits@\!Any measures the proportion of questions for which the predicted answer set overlaps with the gold answer set, i.e., at least one correct answer is predicted:
        \[
        \mathrm{Hits@\!Any} = \frac{1}{|\mathcal{Q}|}\sum_{q\in\mathcal{Q}}\mathbbm{1}[\mathcal{P}_q\cap\mathcal{A}_q\not =\varnothing]\,.
        \]
    \item \textbf{Precision} and \textbf{Recall.}  
        Precision measures the fraction of predicted answers that are correct, while recall measures the fraction of gold answers that are predicted:
        \[
        \mathrm{Precision} = \frac{1}{|\mathcal{Q}|}\sum_{q\in\mathcal{Q}}\frac{|\mathcal{P}_q \cap \mathcal{A}_q|}{|\mathcal{P}_q|}\,,\quad
        \mathrm{Recall} = \frac{1}{|\mathcal{Q}|}\sum_{q\in\mathcal{Q}}\frac{|\mathcal{P}_q \cap \mathcal{A}_q|}{|\mathcal{A}_q|}\,.
        \]
    \item \textbf{F1-score.}
        The F1-score is the harmonic mean of precision and recall, computed per question and averaged across all questions:
        \[
        \mathrm{F1} = \frac{1}{|\mathcal{Q}|}\sum_{q\in\mathcal{Q}}\frac{2\cdot|\mathcal{P}_q \cap \mathcal{A}_q|}{|\mathcal{P}_q|+|\mathcal{A}_q|}\,.
        \]
    \item \textbf{Hard Hits Rate.}
        For each question $q$, the \emph{hard answer} is defined as the selected entity $a_q\in\mathcal{A}_q$ whose supporting triple was intentionally removed from the KG.
        \emph{Hard Hits Rate} (HHR) measures the fraction of correctly answered questions (i.e., Hits@Any) that include the hard answer in predictions:
        \[
        \mathrm{HHR} = \frac{\mathrm{Hits@Hard}}{\mathrm{Hits@Any}}\,,\quad \text{where}\quad \mathrm{Hits@Hard} = \frac{1}{|\mathcal{Q}|}\sum_{q\in\mathcal{Q}}\mathbbm{1}[a_q\in\mathcal{P}_q]\,.
        \]
\end{itemize}


\section{GR-Agent}
\label{sec:approach}
\begin{figure}[t]
    \centering
    \includegraphics[width=\linewidth]{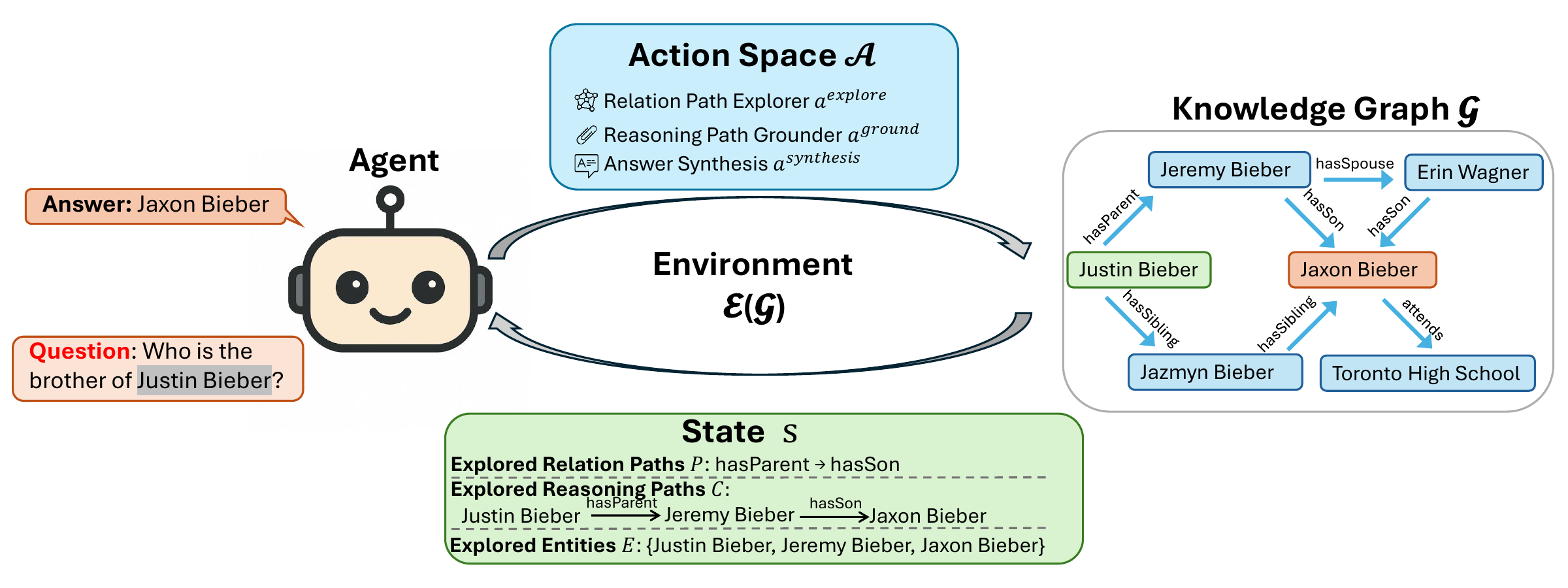}
    \caption{
Overview of our GR-Agent. 
The agent interacts with the environment $\mathcal{E}(\mathcal{G})$, 
which responds to actions by exposing relation paths and entities. 
The agent selects actions from the action space $\mathcal{A}$, 
and each action updates the current state $s$ by updating newly discovered relation paths, grounded reasoning paths, or entities. 
The action space consists of relation-path exploration, path grounding, and answer synthesis.
}

    \label{fig:framework}
    \vspace{-5mm} 
\end{figure}


In this section, we formalize the KGQA task as an agent-environment interaction problem, where is the environment is constructed from a KG $\mathcal{G}$.
Following \citet{sutton1999reinforcement}, we define the environment induced by the KG $\mathcal{G}$ as a tuple
$\mathcal{E}(\mathcal{G}) = (\mathcal{S}, \mathcal{A}, T)$,
where $\mathcal{S}$ is the state space, $\mathcal{A}$ is the action space, and $T$ is  the state transition function.

\subsection{Action Space}
The agent operates over an action space $\mathcal{A}$, which consists of three reasoning tools that enable interaction with the knowledge graph and facilitate question answering.

Basically, the reasoning tools help the agent explore plausible combinations of predicates that can 
form relation paths, ground these  relation paths into reasoning paths, and synthesize 
the gathered reasoning paths to answer the question.

\paragraph{Relation Path Exploration.}
For an entity $e\in\mathcal{E}$ and a hop limit $H\in\mathbb{N}$, the relation path exploration action $a^{\operatorname{explore}}$ generates a set of relation paths of length up to $H$:
\begin{equation}
    a^{\operatorname{explore}}(e, H)\subseteq\mathcal{R}^{\leq H},
\end{equation}
where each relation path $(r_1, \dots, r_k)\in\mathcal{R}^k$ with $k\in [1, H]$ represents a sequence of predicates in $\mathcal{G}$ starting from $e$. 

Concretely, these relation paths are obtained by breadth-first exploration (BFS) \cite{korf1985depth} over the graph starting from the entity $e$. Intuitively, this process collects potential rule bodies (cf. Section~\ref{sec:rule_mining}) that can serve as reasoning skeletons toward the answer.

With this tool, the agent can flexibly decide the exploration range by choosing the hop limit $H$. 
If the current range is insufficient to reach relevant evidence, the agent can iteratively expand it 
(e.g., from $H=1$ to $H=3$) to cover longer relation paths.  

For example, applying the exploration action with an entity \texttt{Justin Bieber} and hop limit $H=2$ on the KG in Figure \ref{fig:framework} yields the following set of relation paths:
\begin{align}\label{eq:relation_path}
    &a^{\operatorname{explore}}(\texttt{Justin Bieber}, 2) = \\
    & 
    \resizebox{0.9\linewidth}{!}{$
    \big\{(\texttt{hasParent}), (\texttt{hasSibling}), (\texttt{hasParent}, \texttt{hasSon}), (\texttt{hasSibling}, \texttt{hasSibling}), (\texttt{hasParent}, \texttt{hasSpouse})\big\}.
    $}
\end{align}
The agent executes this action using the dedicated tool prompt provided in Appendix~\ref{app:exploration_prompt}.

\paragraph{Reasoning Path Grounding.}
Given an entity $e\in\mathcal{E}$, we denote by $P_e(H)$ the set of relation paths returned by the relation path exploration action with hop limit $H$.
The reasoning path grounding action $a^{\operatorname{ground}}$ instantiates each relation path by fixing $e$ as the subject of the first predicate and substituting the remaining variables with entities from the KG, as described in Section \ref{sec:rule_mining}. 
This process produces $a^{\operatorname{ground}}(e, P_e(H)) = (C^{ground}, E^{ground})$, where $C^{ground}$ is the set of grounded triple sequences, which we refer to as \emph{reasoning paths} following the terminology in \citet{linhao2024rog}, and $E^{ground}$ is the set of corresponding entities appearing in these paths, which serve as the \emph{frontier} for further exploration.

With this tool, the agent can instantiate relation paths with concrete entities, 
turning the relation paths into reasoning paths that directly support reasoning, 
via the tool prompt described in Appendix~\ref{app:grounding_prompt}.

For example, grounding the relation paths of topic entity \texttt{Justin Bieber} in Equation~\eqref{eq:relation_path} yields the following set of reasoning paths and frontier entities:
\begin{align}
    &a^{\operatorname{ground}}\big(\texttt{Justin Bieber}, P_{\texttt{Justin Bieber}}(2)\big) = (C^{ground}, E^{ground}), \\
&\notag C^{ground} = \resizebox{0.9\linewidth}{!}{$
    \big\{(\langle\texttt{Justin Bieber}, \texttt{hasParent}, \texttt{Jeremy Bieber}\rangle), 
           (\langle\texttt{Justin Bieber}, \texttt{hasSibling}, \texttt{Jazmyn Bieber}\rangle),
$} \\
    &\notag\quad\quad\quad\quad\quad
    \resizebox{0.9\linewidth}{!}{$
    (\langle\texttt{Justin Bieber}, \texttt{hasParent}, \texttt{Jeremy Bieber}\rangle, 
           \langle\texttt{Jeremy Bieber}, \texttt{hasSon}, \texttt{Jaxon Bieber}\rangle),
               $}\\
    &\notag\quad\quad\quad\quad\quad
    \resizebox{0.9\linewidth}{!}{$
    (\langle\texttt{Justin Bieber}, \texttt{hasSibling}, \texttt{Jazmyn Bieber}\rangle, 
               \langle\texttt{Jazmyn Bieber}, \texttt{hasSibling}, \texttt{Jaxon Bieber}\rangle),
               $}\\
    &\notag\quad \quad\quad\quad\quad
    \resizebox{0.9\linewidth}{!}{$
    (\langle\texttt{Justin Bieber}, \texttt{hasParent}, \texttt{Jeremy Bieber}\rangle, 
               \langle\texttt{Jeremy Bieber}, \texttt{hasSpouse}, \texttt{Erin Wagner}\rangle)\big\}, $}\\
    &\notag E^{ground} = 
    \{\texttt{Jeremy Bieber}, \texttt{Jaxon Bieber}, \texttt{Jazmyn Bieber}, \texttt{Erin Wagner}\}.
\end{align}

\paragraph{Answer Synthesis.}
The answer synthesis action $a^{\operatorname{synthesis}}$ takes as input a set of relevant reasoning paths $C'$, together with the natural language question $q$, and infers the final answers based on this information:
\begin{equation}
    \hat{y}=a^{\operatorname{synthesis}}(C', q).
\end{equation}
The complete set of reasoning paths generated by the executions of the reasoning path grounding action is maintained in the state $s$ and denoted by $C$ (see Section \ref{sec:state_space}). Within this tool, the agent selects the most relevant subset $C'\subseteq C$ to support answer inference, using the tool prompt provided in Appendix \ref{app:answer_synthesis_prompt}.



\subsection{State Space}\label{sec:state_space}
Let $\mathcal{P}=\bigcup_{k=1}^{\infty} \mathcal{R}^k$ denote the set of all possible relation paths, $\mathcal{C}=\bigcup_{k=1}^{\infty} \mathcal{T}^{k}$ denote the set of all possible reasoning paths, where $\mathcal{T}=\mathcal{E}\times\mathcal{R}\times\mathcal{E}$ is the space of all possible triples.
Let $\mathcal{E}$ be the set of all possible entities. 
The state space is then defined as
\begin{equation}
\mathcal{S} = \mathcal{P} \times \mathcal{C} \times \mathcal{E}
\end{equation}
where each state $s=(P,C,E) \in \mathcal{S}$ represent a specific state.

\subsection{State Transition}
The state transition function $T$ defines how the agent's state evolves as it interacts with the environment. Formally,
\begin{equation}
T:\mathcal{S}\times\mathcal{A}\rightarrow\mathcal{S},
\label{eq:state_transition}
\end{equation}
where $\mathcal{S}=\mathcal{P} \times \mathcal{C} \times \mathcal{E}$ is the state space and 
$\mathcal{A}=\{a^{\operatorname{explore}}, a^{\operatorname{ground}}, a^{\operatorname{synthesis}}\}$ is the action space.

At each step, the agent is in a state $s$, which can be viewed as agent's memory,  and selects an action $a\in\mathcal{A}$. Executing $a$ updates the state as follows:
\begin{itemize}
    \item if $a=a^{\operatorname{explore}}$, new relation paths are added to $P$.
    \item if $a=a^{\operatorname{ground}}$, new reasoning paths are added to $C$ and 
          new entities appearing in these paths are added to $E$ (serving as the frontier).
    \item if $a=a^{\operatorname{synthesis}}$, the agent selects a relevant subset 
          $C' \subseteq C$ to infer the final answers and then terminates the interaction.
\end{itemize}

The state transition function \(T\) is realized by the agent 
through a carefully designed system prompt (shown in Appendix~\ref{app:system_prompt}). 
The system prompt lists the available actions $\{a^{\text{explore}}, a^{\text{ground}}, a^{\text{synthesis}}\}$ 
and instructs the model to update the state $s$ until the question can be answered.

\section{Experiments}
\label{sec:exp}




\subsection{Experimental Setup}
\label{sec:exp_setup}

\paragraph{Datasets.} 
We evaluate models on the benchmark datasets introduced in Section~\ref{sec:da_construct}, which instantiate our proposed methodology on Family and FB15k-237. 
For each dataset, we construct both complete and incomplete versions of the KG, where direct supporting triples are removed but sufficient alternative paths are retained to ensure answerability. 
All datasets are split into train/validation/test with an 8:1:1 ratio. 

\paragraph{Evaluation Metrics.} 
Following prior work, we report Hits@Any, precision, recall, and F1. 
In addition, we use the \emph{Hard Hits Rate} (HHR), which specifically measures the proportion of questions for which the model recovers the answer entity that was deliberately removed from the incomplete KG. 
HHR directly reflects the model’s ability to perform multi-hop reasoning under incompleteness. 

\paragraph{Baselines.} 
We evaluate seven representative KG+LLM methods on these benchmarks. 
The \emph{training-based} methods include RoG~\citep{linhao2024rog}, G-Retriever~\citep{he2024g}, and GNN-RAG~\citep{mavromatis2024gnn}. 
The \emph{training-free} methods include PoG~\citep{chen2024pog}, StructGPT~\citep{jiang2023structgpt}, and ToG~\citep{sun2023think}. 
Finally, we compare all baselines against our proposed \textbf{GR-Agent}. 
Training-based methods use the train/validation splits, while training-free methods are evaluated zero-shot; all results are reported on the test split. For training-free models as well as our GR-Agent, we use \emph{GPT-4o-mini} as the underlying LLM.

\subsection{Overall Performance}
\label{sec:overall}

\begin{figure*}[h]
    \centering
    \includegraphics[width=1\linewidth]{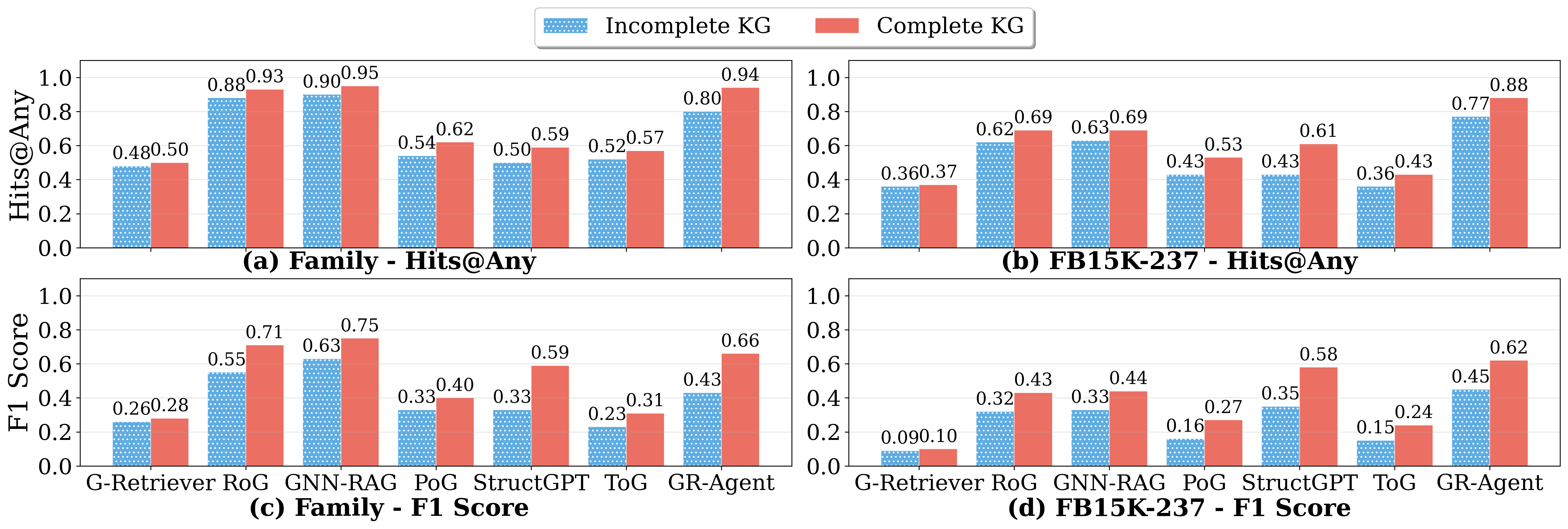}
    \caption{Performance comparison of KG+LLM models under incomplete (blue) and complete (red) KG settings, measured by \textbf{Hits@Any} (top) and \textbf{F1-Score} (bottom).}
    \label{fig:hits_f1}
\end{figure*}

Figure~\ref{fig:hits_f1} reports Hits@Any and F1-scores across all methods and datasets. 
In most cases, \textbf{both metrics drop noticeably when moving from the complete to the incomplete KG setting}, highlighting the challenge posed by missing direct supporting triples. 
Precision and recall follow a similar trend (see Appendix~\ref{sec:app_exp}).
Overall, training-based methods (e.g., RoG and GNN-RAG) achieve stronger performance than most non-trained approaches, while G-Retriever remains consistently low due to its reliance on noisy textual similarity retrieval rather than reasoning. 

\textbf{Our GR-Agent achieves the best results among training-free methods on both datasets} and exhibits a smaller performance drop compared to existing non-trained baselines, demonstrating stronger robustness to incompleteness. 
On FB15k-237, GR-Agent further surpasses training-based methods such as RoG and GNN-RAG, achieving the best overall balance between Hits@Any and F1-score. 
These results confirm the effectiveness of path-centric reasoning for improving robustness under KG incompleteness.

\subsection{Impact of Removing Supporting Triples}
\label{sec:remov_direct}

\begin{figure*}[t] 
    \centering
    \begin{subfigure}{0.48\textwidth}
        \centering
        \includegraphics[width=\linewidth]{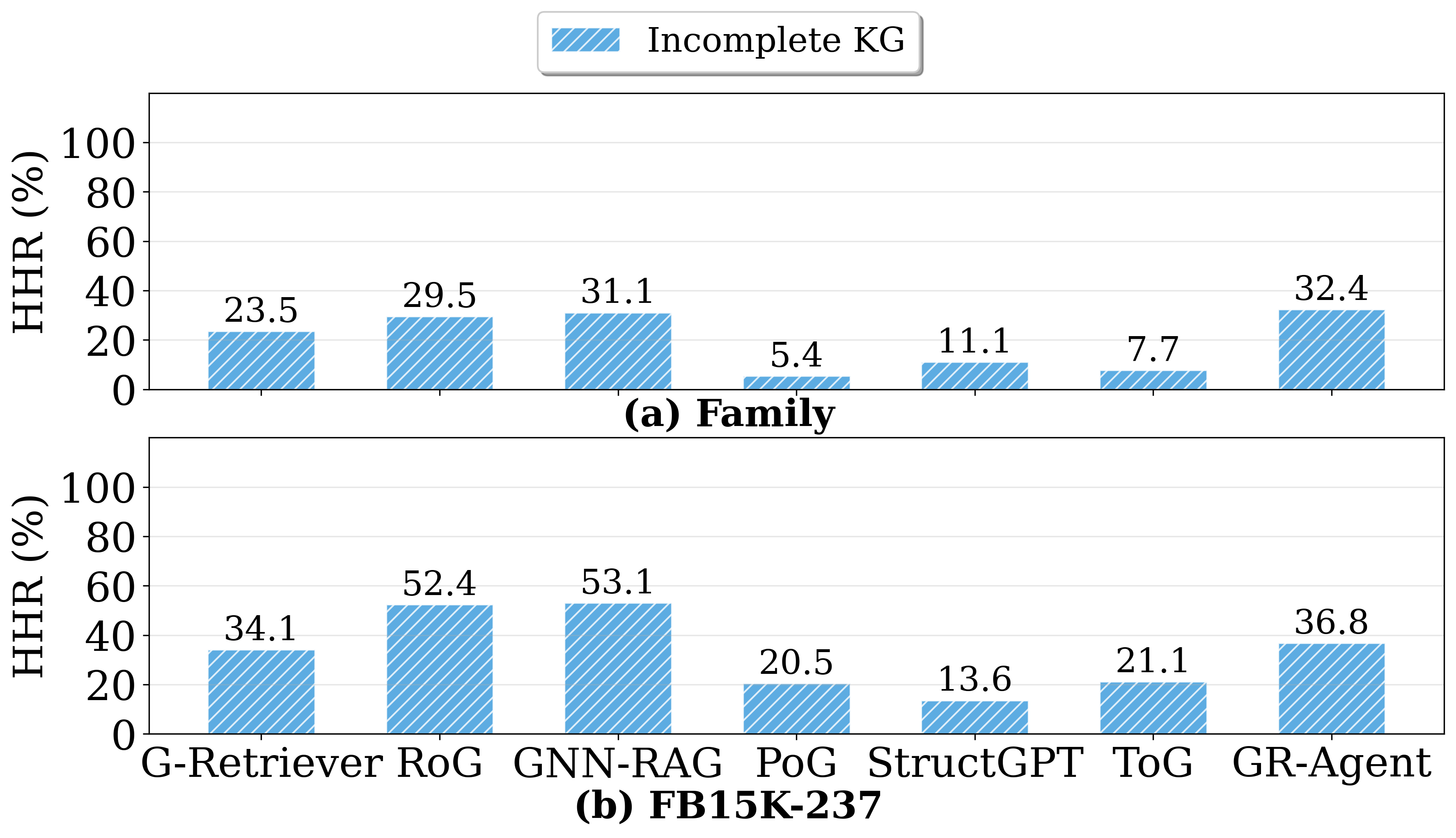}
        \caption{HHR under different KG settings.}
        \label{fig:hard_hits}
    \end{subfigure}
    \hfill
    \begin{subfigure}{0.48\textwidth}
        \centering
        \includegraphics[width=\linewidth]{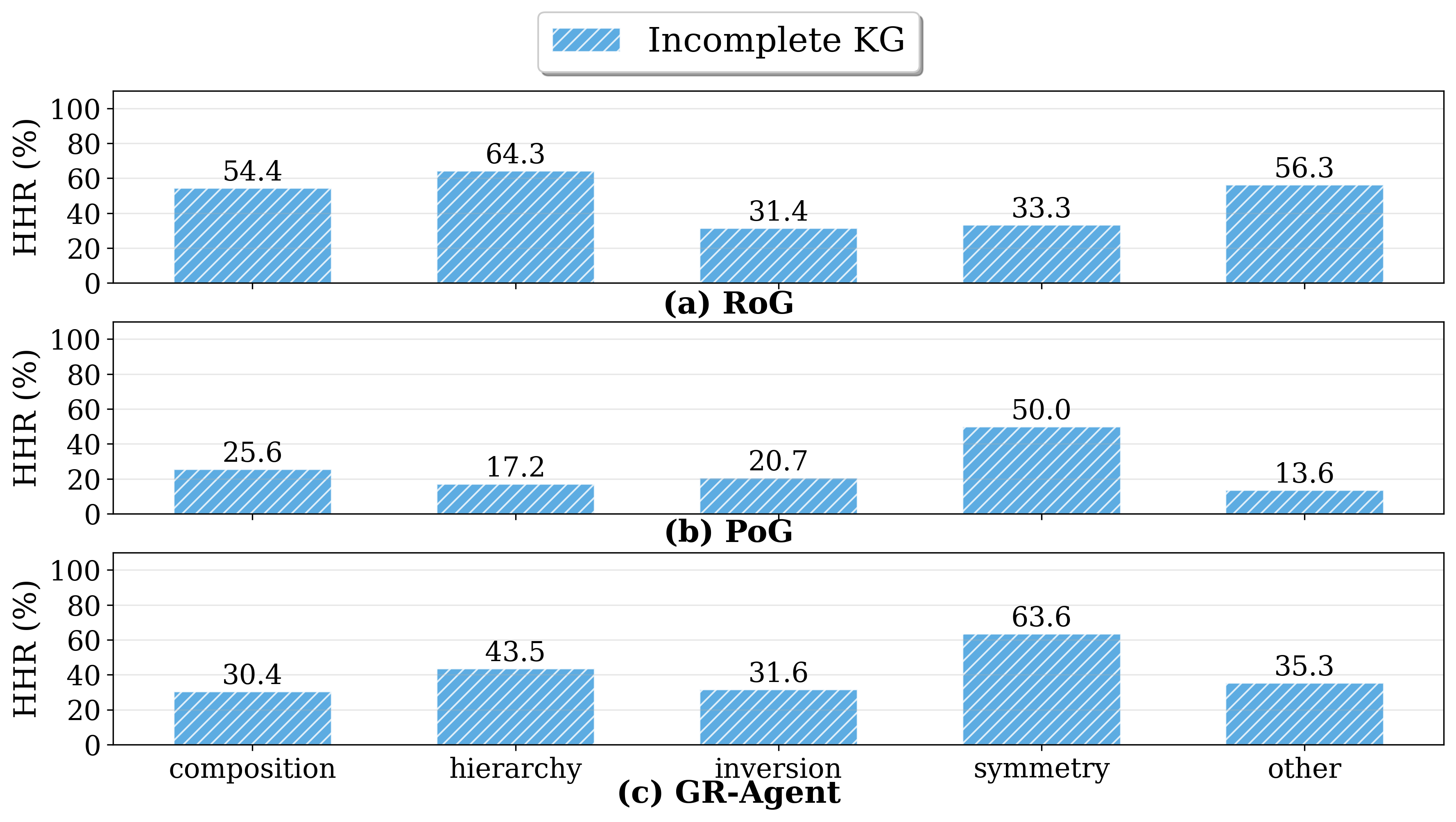}
        \caption{HHR across rule types on FB15k-237.}
        \label{fig:hard_hits_rog_pog}
    \end{subfigure}
    
    \caption{Comparative analysis: (a) performance under different KG settings; (b) performance across rule types.}
    \label{fig:combined_hhr}
        \vspace{-5mm} 
\end{figure*}



To examine the impact of removing direct supporting triples, we report the Hard Hits Rate (HHR) under 
the \emph{incomplete KG} setting( Figure~\ref{fig:hard_hits}).
Recall that HHR is defined as the fraction of 
correctly answered questions (Hits@Any) in which the hard answer whose direct supporting triples 
were deliberately removed is also recovered.
It thus provides a 
meaningful indicator of reasoning ability when direct supporting triples are absent. 

Across both datasets, the  HHR remains modest, indicating that even when models 
successfully answer a question, they often fail to recover the missing hard answer. This 
underscores the difficulty of reconstructing facts through alternative reasoning paths. 
Training-based methods (e.g., RoG and GNN-RAG) generally achieve higher HHR than non-trained 
methods (e.g., PoG and ToG), suggesting that exposure to missing-evidence scenarios during 
training improves generalization over multi-hop reasoning paths.

In contrast, among non-trained methods, our GR-Agent achieves the highest HHR under incomplete KGs, 
indicating substantially stronger robustness to missing facts. While other training-free approaches 
degrade sharply once direct support triples are absent, GR-Agent is able to identify and leverage alternative reasoning paths.
This demonstrates that \textbf{GR-Agent achieves the strongest and most generalizable reasoning ability 
under incomplete KGs among non-trained approaches, while remaining comparable to training-based methods}.

\subsection{Fine-Grained Analysis by Rule Type}
\label{sec:rule_type}

To gain deeper insights into reasoning behavior, we break down HHR by rule type on FB15k-237, 
comparing RoG, PoG, and our GR-Agent (Figure~\ref{fig:hard_hits_rog_pog}). 

Overall, RoG exhibits the highest robustness across most rule types, reflecting the benefit of 
training-based adaptation to missing evidence. PoG, in contrast, struggles on all categories 
except symmetry, where it achieves 50.0\% HHR. This is consistent with the observation that 
symmetric relations (e.g., \texttt{sibling}) are easier for an LLM to handle, since reversing 
arguments does not require complex multi-hop reasoning. However, PoG’s poor performance on 
hierarchy and ``other'' rules highlights its strong reliance on shallow retrieval. 

Our proposed GR-Agent consistently outperforms PoG and narrows the gap with RoG. Notably, it reaches 
63.6\% HHR on symmetry and 43.5\% on hierarchy rules, showing robustness across both simple and 
structurally complex patterns. While still below RoG on composition-heavy cases, GR-Agent surpasses 
PoG by large margins across all rule types, demonstrating that adaptive path exploration enables 
stronger generalization even without training.

\subsection{Case Study}
\label{sec:case}

To better understand the limitations of current KG+LLM methods, we analyze representative failure cases from our benchmarks. Table~\ref{tab:case-study} shows two illustrative examples, each highlighting a distinct failure pattern: (1) insufficient multi-hop retrieval and (2) reasoning failure despite correct retrieval.



\paragraph{(1) Example 1: Insufficient Multi-hop Retrieval.}
Answering ``What is the country of the administrative division Calvados?'' requires a three-hop path.
Existing retrievers instead return shorter partial paths,
which stall at intermediate nodes. Lacking the final hop, the generator hallucinates answers like Spain. By contrast, the GR-Agent Trace successfully explores the full relation path, grounds it, and synthesizes the correct answer.  

\paragraph{(2) Example 2: Reasoning Failure.}
Here, the retriever does return the correct supporting triple \texttt{spouse(Ian Holm, Penelope Wilton)}, which should enable inferring the reverse direction. Yet the generator outputs Marriage, distracted by unrelated \texttt{typeOfUnion} paths. This shows that even with correct retrieval, models may fail to prioritize relevant evidence during generation. The GR-Agent Trace instead focuses on the relevant relation path, grounds it correctly, and outputs the right answer.

\begin{table*}[h]
\centering
\resizebox{\linewidth}{!}{
\begin{tabular}{@{}l@{}}
\toprule
\textbf{Example 1:} \textit{Question:} What is the country of the administrative division \textbf{Calvados}? 
\quad \textit{Answer:} [France] \\
\textit{Alternative Path:} \texttt{\textcolor{green!60!black}{capital(Calvados, Caen)} $\wedge$ \textcolor{green!60!black}{capitalOf(Caen, Calvados)} $\wedge$ \textcolor{green!60!black}{contains(France, Calvados)}} 
$\Rightarrow$ \texttt{country(Calvados, France)} \\
--- \\
\textit{Prediction:} [\textcolor{red}{Spain}] \\
\textit{Retrieved Paths:} 
\quad \texttt{administrativeParent(Calvados, LowerNormandy)} $\wedge$ \texttt{contains(Calvados, Caen)} \\
--- \\
\textbf{GR-Agent Trace:} \\
\quad \textit{(1) Relation Path Exploration:} \texttt{capital $\rightarrow$ capitalOf, contains} \\
\quad \textit{(2) Reasoning Path Grounding:} \texttt{capital(Calvados, Caen)} $\wedge$ \texttt{capitalOf(Caen, Calvados)}; \texttt{contains(France, Calvados)} 
\quad \textit{(3) Answer Synthesis:} [\textcolor{green!60!black}{France}] \\
\midrule

\textbf{Example 2:} \textit{Question:} Who is \textbf{Ian Holm}'s spouse? \quad \textit{Answer:} [Penelope Wilton] \\
\textit{Alternative Paths:} \texttt{\textcolor{green!60!black}{spouse(Ian Holm, Penelope Wilton)} $\Rightarrow$ spouse(Penelope Wilton, Ian Holm)} \\
--- \\
\textit{Prediction:} [\textcolor{red}{Marriage}] \\
\textit{Retrieved Paths:} 
\quad \texttt{\textcolor{green!60!black}{spouse(Ian Holm, Penelope Wilton)}} \\
\quad \texttt{awardNominee(Ian Holm, Cate Blanchett) $\wedge$ typeOfUnion(Cate Blanchett, \textcolor{red}{Marriage})} \\
\quad \texttt{awardNominee(Ian Holm, Kate Beckinsale) $\wedge$ typeOfUnion(Kate Beckinsale, Domestic Partnership)} \\
--- \\
\textbf{GR-Agent Trace:} \\
\quad \textit{(1) Relation Path Exploration:} \texttt{spouse} 
\quad \textit{(2) Reasoning Path Grounding:} \texttt{spouse(Ian Holm, Penelope Wilton)} 
\quad \textit{(3) Answer Synthesis:} [\textcolor{green!60!black}{Penelope Wilton}] \\
\midrule
\bottomrule
\end{tabular}
}
\caption{Case studies comparing baseline failures with GR-Agent. 
Unlike KG+LLM models that retrieve incomplete multi-hop paths or follow noisy evidence, GR-Agent performs multiple actions to derive the correct answer. (\textcolor{green!60!black}{green} = expected paths, \textcolor{red}{red} = incorrect predictions).}
\label{tab:case-study}
    \vspace{-5mm} 

\end{table*}

\section{Discussion and Conclusion}

In this work, we studied the challenge of reasoning under incomplete KGs, where direct supporting triples are missing and answers must be inferred through alternative paths. To address this, we proposed \textbf{GR-Agent}, a training-free adaptive graph reasoning agent  that explores relation paths, grounds them into reasoning paths, and synthesizes answers with the help of a lightweight planner and evolving memory. Alongside the method, we introduce a general construction methodology for simulating incomplete KGs while ensuring answerability: direct supporting triples are removed, but sufficient alternative paths are preserved. We instantiate this paradigm on Family and FB15k-237, providing two benchmarks for systematic assessment of reasoning ability under incompleteness.  

Experiments show that GR-Agent consistently outperforms other training-free methods and achieves performance comparable to training-based systems.
These results suggest that path-centric exploration is an effective way to recover missing knowledge in real-world KGs. This work is ongoing. In the future, we plan to extend GR-Agent with reinforcement learning to optimize tool usage and further improve reasoning efficiency. 


\section{Acknowledgements}
The authors thank the International Max Planck Research School for Intelligent Systems (IMPRS-IS) for supporting Yuqicheng Zhu and Hongkuan Zhou. 
The work was partially supported by EU Projects Graph Massivizer (GA 101093202), enRichMyData (GA 101070284), SMARTY (GA 101140087), and the EPSRC project OntoEm (EP/Y017706/1).




\bibliographystyle{plainnat} 
\bibliography{custom}
\newpage
\appendix

\section{Details of Rule Mining}
\label{app:rule_mining}
\textsc{AMIE3} is a widely used rule mining system designed to operate efficiently over large-scale KGs. 
A logical rule discovered by \textsc{AMIE3} has the following form (\emph{Horn rules}~\cite{horn1951sentences}):
\begin{equation*}
B_1 \wedge B_2 \wedge \dots \wedge B_n \Rightarrow H\,,
\label{eq:horn}
\end{equation*}
where each item is called an \emph{atom}, a binary relation of the form $r(X,Y)$, in which $r$ is a predicate and $X,Y$ are variables. The left-hand side of the rule is a conjunction of \emph{body atoms}, denoted as $\mathbf{B}=B_1\wedge\dots\wedge B_n$, and the right-hand side is the \emph{head atom} $H$. Intuitively, a rule expresses that if the body $\mathbf{B}$ holds, then the head $H$ is likely to hold as well.

A \emph{substitution} $\sigma$ maps every variable occurring in an atom to a entity that exists in $\mathcal{G}$.
For example applying $\sigma=\{X\mapsto \texttt{Justin}, Y\mapsto \texttt{Jaxon}\}$ to the atom \texttt{hasSibling(X,Y)} yields the grounded fact \texttt{hasSibling(Justin,Jaxon)}. A \emph{grounding} of a rule $\mathbf{B}\Rightarrow H$ is 
\begin{equation*}
    \sigma(B_1)\wedge\dots\sigma(B_n)\Rightarrow\sigma(H)\,.
\end{equation*}

\paragraph{Quality Measure.} \textsc{AMIE3} uses the following metrics to measure the quality of a rule:
\begin{itemize}
    \item \textbf{Support.} The \emph{support} of a rule is defined as the number its groundings for which all grounded facts are observed in the KG:
    \begin{align*}
        &support(\mathbf{B}\Rightarrow H) = \\
        &|\{\sigma(H)\mid \forall i,\sigma(B_i)\in\mathcal{G}\wedge\sigma(H)\in\mathcal{G}\}|\,.
    \end{align*}
 
    \item \textbf{Head coverage.} \emph{Head coverage (hc)} measures the proportion of observed head groundings in the KG that are successfully explained by the rule. It is defined as the ratio of the rule’s support to the number of head groundings in the KG:
    \begin{equation*}
        hc(\mathbf{B}\Rightarrow H) = \frac{support(\mathbf{B}\Rightarrow H)}{|\{\sigma\mid\sigma(H)\in\mathcal{G}\}|}\,.
    \end{equation*}
    \item \textbf{Confidence.} \emph{Confidence} measures the proportion of body groundings that also lead to the head being observed in the KG. It is defined as the ratio of the rule's support to the number body groundings in the KG:
    \begin{equation*}
        confidence(\mathbf{B}\Rightarrow H) = \frac{support(\mathbf{B}\Rightarrow H)}{|\{\sigma\mid\sigma(\mathbf{B})\in\mathcal{G}\}|}\,.
    \end{equation*}
\end{itemize}

We retain only rules with high confidence and sufficient support, filtering out noisy or spurious patterns. Specifically, we run \textsc{AMIE3} with a confidence threshold of 0.3, a head coverage threshold of 0.1, and a maximum rule length of 4. \textsc{AMIE3} incrementally generates candidate rules via breadth-first refinement~\cite{lajus2020fast} and evaluates them using confidence and head coverage; only those meeting the specified thresholds are retained. Additional details on the rule generation and filtering process are provided in Appendix~\ref{app:rule_mining}.

\subsection{AMIE3 Candidate Rule Refinement}
Refinement is carried out using a set of operators that generate new candidate rules:
\begin{itemize}
    \item Dangling atoms, which introduce a new variable connected to an existing one;
    \item Closing atoms, which connect two existing variables;
    \item Instantiated atoms, which introduce a constant and connect it to an existing variable.
\end{itemize}

AMIE3 generate candidate rules by a refinement process using a classical breadth-first search~\cite{lajus2020fast}. It begins with rules that contain only a head atom (e.g. $\Rightarrow \texttt{hasSibling}(X,Y)$) and refines them by adding atoms to the body. For example, it may generate the refined rule:
\begin{align*}
    \texttt{hasParent}(X,Z)&\wedge \texttt{hasChild}(Z,Y)\\
    &\Rightarrow \texttt{hasSibling}(X,Y)\,.
\end{align*}
This refinement step connects existing variables and introduces new ones, gradually building meaningful patterns. 

\subsection{AMIE3 Hyperparameter Settings}
\label{sec:amie3_hyper}

We use AMIE3 with a confidence threshold of 0.3 and a PCA confidence threshold $\theta_{\text{PCA}}$ of 0.4 for both datasets. The maximum rule length is set to 3 for \textbf{Family} to avoid overly complex patterns, and 4 for \textbf{FB15k-237} to allow richer rules. See Appendix~\ref{app:pca_conf} for the definition of PCA confidence.

\section{Properties of Horn Rules Mined by AMIE3}
\label{app:amie3_horn_properties}

AMIE3 mines logical rules from knowledge graphs in the form of (Horn) rules:
\[
B_1 \wedge B_2 \wedge \cdots \wedge B_n \implies H
\]
where $B_i$ and $H$ are atoms of the form $r(X, Y)$. To ensure interpretability and practical utility, AMIE3 imposes the following structural properties on all mined rules:

\begin{itemize}
    \item \textbf{Connectedness:} All atoms in the rule are transitively connected via shared variables or entities. This prevents rules with independent, unrelated facts (e.g., $\texttt{diedIn}(x, y) \implies \texttt{wasBornIn}(w, z)$). Two atoms are connected if they share a variable or entity; a rule is connected if every atom is connected transitively to every other atom.
    \item \textbf{Closedness:} Every variable in the rule appears at least twice (i.e., in at least two atoms). This avoids rules that merely predict the existence of some fact without specifying how it relates to the body, such as $\texttt{diedIn}(x, y) \implies \exists z: \texttt{wasBornIn}(x, z)$.
    \item \textbf{Safety:} All variables in the head atom also appear in at least one body atom. This ensures that the rule's predictions are grounded by the body atoms and avoids uninstantiated variables in the conclusion.
\end{itemize}

These restrictions are widely adopted in KG rule mining~\citep{galarraga2015fast,lajus2020fast} to guarantee that discovered rules are logically well-formed and meaningful for downstream reasoning tasks.

\subsection{PCA Confidence}
\label{app:pca_conf}
To understand the concept of rule mining better for the reader we simplified notation of confidence in main body.
Note \textsc{AMIE3} also supports a more optimistic confidence metric known as \emph{PCA confidence}, which adjusts standard confidence to account for incompleteness in the KG. 

\textbf{Motivation.}
Standard confidence for a rule is defined as the proportion of its correct predictions among all possible predictions suggested by the rule. However, this metric is known to be pessimistic for knowledge graphs, which are typically incomplete: many missing triples may be true but unobserved, unfairly penalizing a rule's apparent reliability.

\textbf{Definition.}
To address this, AMIE3 introduces \emph{PCA confidence} (Partial Completeness Assumption confidence)~\citep{galarraga2015fast}, an optimistic variant that partially compensates for KG incompleteness. Given a rule of the form
\[
B_1 \wedge \cdots \wedge B_n \implies r(x, y)
\]
the \textbf{standard confidence} is
\[
\mathrm{conf}(R) = \frac{|\{(x, y): B_1 \wedge \cdots \wedge B_n \wedge r(x, y)\}|}
{|\{(x, y): B_1 \wedge \cdots \wedge B_n\}|}
\]
where the denominator counts all predictions the rule could possibly make, and the numerator counts those that are actually present in the KG.

\textbf{PCA confidence} modifies the denominator to include only those $(x, y)$ pairs for which at least one $r(x, y')$ triple is known for the subject $x$. That is, the rule is only penalized for predictions about entities for which we have observed at least some information about the target relation. Formally,
\begin{align*}
    &\mathrm{conf_{PCA}}(R) = \\
&\frac{|\{(x, y): B_1 \wedge \cdots \wedge B_n \wedge r(x, y)\}|}
{|\{(x, y): B_1 \wedge \cdots \wedge B_n \wedge \exists y': r(x, y')\}|}
\end{align*}

Here, the denominator sums only over those $x$ for which some $y'$ exists such that $r(x, y')$ is observed in the KG.

\textbf{Intuition.}
This approach assumes that, for any entity $x$ for which at least one fact $r(x, y')$ is known, the KG is "locally complete" with respect to $r$ for $x$, so if the rule predicts other $r(x, y)$ facts for $x$, and they are missing, we treat them as truly missing (i.e., as counterexamples to the rule). But for entities where no $r(x, y)$ fact is observed at all, the rule is not penalized for predicting additional facts.

\textbf{Comparison.}
PCA confidence thus provides a more optimistic and fairer assessment of a rule's precision in the presence of incomplete data. It is widely adopted in KG rule mining, and is the default metric for filtering and ranking rules in AMIE3.

For further details, see~\citep{galarraga2015fast}.

\subsection{Rule Mining Procedure}

\begin{algorithm}[h]
\caption{AMIE3}
\label{algo:AMIE}
    \begin{algorithmic}[1]
        \Require Knowledge graph $\mathcal{G}$, maximum rule length $l$, PCA confidence threshold $\theta_{PCA}$, and head coverage threshold $\theta_{hc}$.
        \Ensure Set of mined rules $\mathcal{R}$.
        \State $q \gets$ all rules of the form $\top \Rightarrow r(X, Y)$
        \State $\mathcal{R} \gets \emptyset$
        \While{$q$ is not empty}
            \State $R \gets$ $q$.dequeue()
            \If{$\texttt{SatisfiesRuleCriteria}(R)$}
                \State $\mathcal{R} \gets \mathcal{R} \cup \{R\}$
            \EndIf
            \If{$\texttt{len}(R) < l$ and $\theta_{PCA}(R) < 1.0$}

                \ForAll{$R_c \in \texttt{refine}(R)$}
                    \If{$hc(R_c) \ge \theta_{hc}$ and $R_c \notin q$}
                        \State $q$.enqueue($R_c$)
                    \EndIf
                \EndFor
            \EndIf
        \EndWhile
        \State \Return $\mathcal{R}$
    \end{algorithmic}
\end{algorithm}

\textsc{AMIE3} generate candidate rules by a refinement process using a classical breadth-first search~\cite{lajus2020fast}.
Algorithm~\ref{algo:AMIE} summarizes the rule mining process of \textsc{AMIE3}. 
The algorithm starts with an initial set of rules that contain only a head atom (i.e. $\top \Rightarrow r(X, Y)$, where $\top$ denotes an empty body) and maintains a queue of rule candidates (Line 1). 
At each step, \textsc{AMIE3} dequeues a rule $R$ from the queue and evaluates whether it satisfies three criteria (Line 5):
\begin{itemize}
    \item the rule is \emph{closed} (i.e., all variables in at least two atoms),
    \item its PCA confidence is higher than $\theta_{PCA}$,
    \item its PCA confidence is higher than the confidence of all previously mined rules with the
same head atom as $R$ and a subset of its body atoms.
\end{itemize} 
If these conditions are met, the rule is added to the final output set $\mathcal{R}$. 

If $R$ has fewer than $l$ atoms and its confidence can still be improved (Line 8), \textsc{AMIE3} applies a \texttt{refine} operator (Line 9) that generates new candidate rules by adding a body atom (details in Appendix~\ref{app:rule_mining}). 
Refined rules are added to the queue only if they have sufficient head coverage (Line 11) and have not already been explored.
This process continues until the queue is empty, at which point all high-quality rules satisfying the specified constraints have been discovered.

\section{Details of Dataset Generation}
\label{app:downsample}
KGs typically exhibit a "long-tail" distribution, where a small number of entities participate in a disproportionately large number of triples, while the majority appear only infrequently ~\citep{mohamed2020popularity, chen2023knowledge}.
This imbalance can cause many generated questions to share the same answer entity, leading to biased evaluation.

To reduce answer distribution bias, we apply frequency-based downsampling to the generated questions $\mathcal{Q}$, yielding a more balanced subset $\mathcal{Q}' \subseteq \mathcal{Q}$. 
The procedure is summarized in Algorithm~\ref{alg:downsample}, which is provided in Appendix~\ref{app:downsample}. 
Specifically, for each answer entity $a$, we retain at most $\tau \cdot |\mathcal{Q}|$ questions if $a$ exceeds the frequency threshold $\tau$; otherwise, all associated questions are kept.

\begin{algorithm}[h]
\caption{Downsampling Procedure}
\label{alg:downsample}
\begin{algorithmic}[1]
\Require Question set $\mathcal{Q}$; threshold $\tau \in (0, 1]$
\Ensure Balanced subset $\mathcal{Q}' \subseteq \mathcal{Q}$

\State Let $\mathcal{A} \gets$ set of unique answer entities in $\mathcal{Q}$
\State $\mathcal{Q}' \gets \emptyset$
\ForAll{$a \in \mathcal{A}$}
    \State $\mathcal{Q}_a \gets \{q \in \mathcal{Q} \mid \texttt{answer}(q) = a\}$
    \If{$|\mathcal{Q}_a| > \tau \cdot |\mathcal{Q}|$}
        \State Randomly sample $\mathcal{S}_a \subset \mathcal{Q}_a$ 
        \State of size $\lfloor \tau \cdot |\mathcal{Q}| \rfloor$
    \Else
        \State $\mathcal{S}_a \gets \mathcal{Q}_a$
    \EndIf
    \State $\mathcal{Q}' \gets \mathcal{Q}' \cup \mathcal{S}_a$
\EndFor
\State \Return $\mathcal{Q}'$
\end{algorithmic}
\end{algorithm}

\subsection{Benchmark Construction Code and Data}

We release the source code for benchmark construction, along with the Family and FB15k-237 benchmark datasets, at \url{https://anonymous.4open.science/r/INCK-EA16}.

\section{Additional Results of the experiment}
\begin{figure*}[h]
    \centering
    \includegraphics[width=1\linewidth]{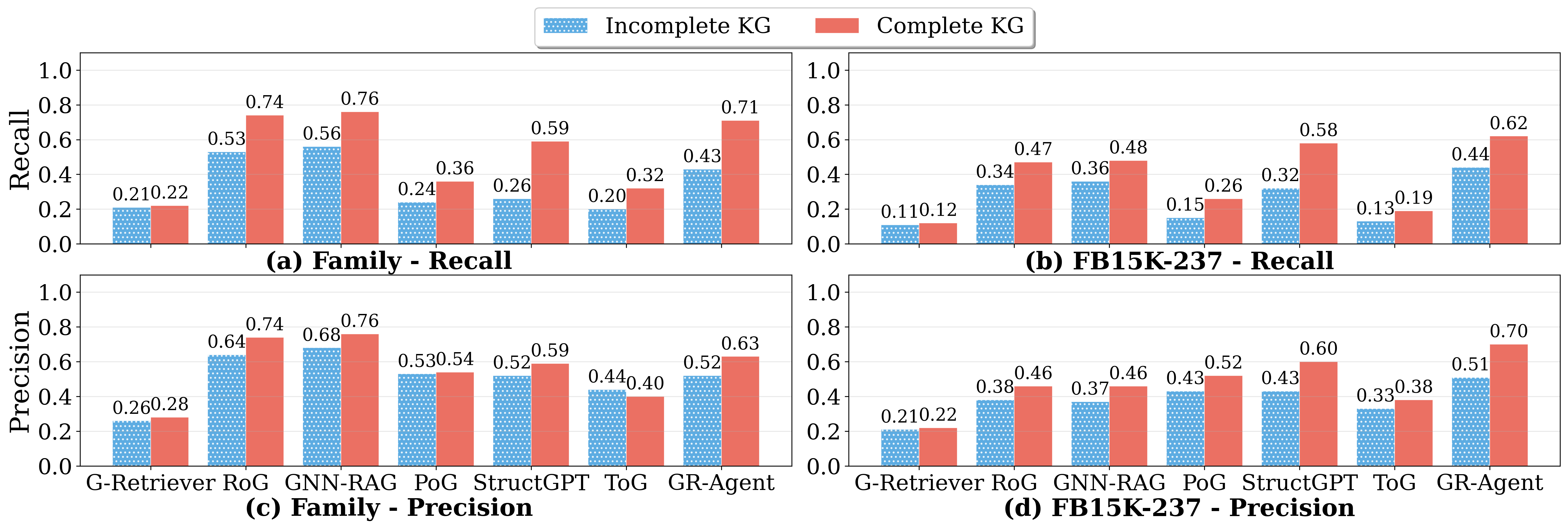}
    \caption{Performance comparison of KG-RAG models under incomplete (blue) and complete (red) KG settings, measured by \textbf{Recall} (top) and \textbf{Precision} (bottom).}
    \label{fig:precision_recall}
\end{figure*}

Figure~\ref{fig:precision_recall} presents the recall and precision of all evaluated KG-RAG models on the constructed benchmarks.

\label{sec:app_exp}

\section{Prompt Template}\label{app:generation_prompt}
Prompt for generating questions from triples:
\begin{lstlisting}[style=kgprompt]
You are an expert in knowledge graph question generation.

Given:
Removed Triple: ({entity_h}, {predicate_T}, {entity_t})
Question Entity: {topic_entity}
Answer Entity: {answer_entity}

Write a clear, natural-language question that asks for the Answer Entity, using the given predicate and Topic Entity.

Requirements:
- Express the predicate {predicate_T} naturally (paraphrasing allowed, but preserve core meaning; e.g., "wife_of" -> "wife").
- Mention the Topic Entity {topic_entity}.
- The answer should be the Answer Entity {answer_entity}.
- Do not mention the Answer Entity {answer_entity} in the question.
- Do not ask a yes/no question.
- Output only the question as plain text.

Example:
Removed Triple: ("Alice", "wife_of", "Carol")
Question Entity: Carol
Answer Entity: Alice

Output:
Who is Carol's wife?

Now, generate the question for:
Removed Triple: ({entity_h}, {predicate_T}, {entity_t})
Question Entity: {topic_entity}
Answer Entity: {answer_entity}
\end{lstlisting}

To ensure reproducibility and mitigate randomness in LLM outputs~\cite{potyka2024robust}, we set the generation temperature to 0 in all experiments.

\section{Tool Prompt for Relation Path Exploration}\label{app:exploration_prompt}
\begin{lstlisting}[style=kgprompt]
Mine relation paths from an entity to discover reasoning patterns.


This tool returns ALL paths from 1-hop up to max_hops combined in one list.

CRITICAL USAGE:
- Select the starting entity strategically.
- Use small hop limits for efficient exploration, and increase gradually 
   if evidence is insufficient.
- Collected relation paths will serve as potential reasoning skeletons 
  for grounding and synthesis.

Args:
    entity: Starting entity for exploration 
    max_hops: Maximum path length 
    

Returns:
    Combined relation paths represented as strings, e.g.,
    ['rel1', 'rel1 -> rel4', 'rel2 -> rel1']
\end{lstlisting}

\section{Tool Prompt for Reasoning Path Grounding}\label{app:grounding_prompt}
\begin{lstlisting}[style=kgprompt]
Ground relation paths to find concrete entity sequences that answer the question.


This tool finds actual entity sequences that follow the selected patterns, providing
concrete evidence for reasoning.


Args:
    entity: Starting entity for grounding 
    relation_paths: Selected relation path strings from relation_path_match
   

Returns:
    Grounded path descriptions with entity sequences and evidence triples
\end{lstlisting}

\section{Tool Prompt for Answer Synthesis}\label{app:answer_synthesis_prompt}
\begin{lstlisting}[style=kgprompt]
Complete the knowledge graph exploration when reasoning paths are sufficient 
to answer the question.

The agent should return the final answer based on reasoning over the discovered 
reasoning paths to terminate the exploration.

CRITICAL QUESTION UNDERSTANDING:
- Carefully analyze what the question is asking for.
- Extract answer entities that directly answer the question, not related but 
  irrelevant entities
- Select only reasoning paths that lead to the correct entity type being asked for

The explored_reasoning_paths are formatted as strings containing the grounded 
path evidence:
     Evidence: <supporting_reasoning_paths>


The agent should focus on answering the original question using reasoning over 
these paths. Use the reasoning paths to infer the correct answer entities 
through pattern matching and evidence analysis.



Reasoning strategies:
- Direct matches: triples that directly answer the query
- Fuzzy matches: similar relation/entity names that approximately match the target
- Inverse relationships: if you find "A relation B", consider "B inverse_relation A"
- Chain reasoning: use patterns like "A rel1 B" + "B rel2 C" to infer "A answers C"
- Evidence stacking: multiple consistent triples together provide sufficient evidence

CRITICAL: Among the explored reasoning paths, only a subset can actually answer 
the question. The agent should carefully select the most reasonable and reliable 
subset as supporting reasoning paths. Note that entities appearing in reasoning 
paths as intermediate steps may not be answer entities. Use them to infer the 
answer while excluding false evidence. 

Args:
    explored_reasoning_paths (list[str]): Grounded reasoning paths that support the final answer
    answer_entities (list[str]): Final answer entity IDs only

Returns:
    dict[str, Any]: Final results with answers and reasoning path evidence
\end{lstlisting}

\section{System Prompt}\label{app:system_prompt}
Default system prompt for knowledge graph reasoning:
\begin{lstlisting}[style=kgprompt]
You are a helpful assistant that answers queries by exploring a knowledge 
graph using advanced path-based reasoning.

Available tools:
- relation_path_mining: Discover all possible relation paths around an entity to find reasoning patterns.
- path_grounding: Instantiate selected relation paths with concrete entities 
  to find actual reasoning chains.
- complete_task: Finalize the answer using reasoning paths as evidence.

The toolkit maintains an evidence store of discovered entities and triples. 
Use the tools iteratively--discover, select, extract--then finalize when ready 
with concrete entity answers.

Important context:
- Real-world knowledge graphs are always incomplete. Do NOT expect to always 
  find a direct triple that answers the question.
- Instead, you must rely on indirect evidence, combining multiple facts and 
  relation paths. If no direct edge exists, reason over intermediate nodes and 
  multi-hop chains to imply the answer.
- Avoid finalizing prematurely if only partial evidence is present; keep 
  exploring relation paths to assemble a reasoning chain.

Key principles:
- Focus on RELATION PATHS as reasoning patterns, not individual triples
- Multi-hop reasoning is essential -- explore 1-hop, 2-hop, and 3-hop patterns
- Select paths strategically based on semantic relevance to the question
- Ground selected paths to get concrete evidence chains
- A reasoning path can connect topic entity and answer entity through 
  intermediate entities
- Look for both direct relations and inverse relations
- Use path grounding results as structured evidence for your final answer
\end{lstlisting}

\section{Detailed Evaluation Settings}\label{app:eval_setting}
All evaluated models are required to produce their predictions as a \emph{list of answers}, but in practice, the model output is often a raw string $P_{\mathrm{str}}$ (e.g., \texttt{"Paris, London"} or \texttt{"Paris  London"}). To obtain a set-valued prediction suitable for evaluation, we first apply a splitting function $\mathrm{split}(P_{\mathrm{str}})$, which splits the raw string into a list of answer strings $P = [p_1, p_2, \ldots, p_n]$ using delimiters such as commas, spaces, or newlines as appropriate.

We then define a normalization function $\mathrm{norm}(\cdot)$, which converts each answer string to lowercase, removes articles (\texttt{a}, \texttt{an}, \texttt{the}), punctuation, and extra whitespace, and eliminates the special token \texttt{<pad>} if present. The final prediction set is then defined as $\mathcal{P} = \{\mathrm{norm}(p) \mid p \in P\}$, i.e., the set of unique normalized predictions. The same normalization is applied to each gold answer in the list $A$ to obtain the set $\mathcal{A}$.

All evaluation metrics are computed based on the resulting sets of normalized predictions $\mathcal{P}$ and gold answers $\mathcal{A}$.

\begin{algorithm}
\caption{Output Processing}
\label{alg:normalize}
\begin{algorithmic}[1]
\Require Model output string $P_{\mathrm{str}}$, gold answer list $A$
\Ensure Normalized prediction set $\mathcal{P}$, normalized gold set $\mathcal{A}$
\State $P \gets \mathrm{split}(P_{\mathrm{str}})$
\State $\mathcal{P} \gets \{\, \mathrm{norm}(p) \mid p \in P \,\}$ 
\State $\mathcal{A} \gets \{\, \mathrm{norm}(a) \mid a \in A \,\}$ 
\State \Return $\mathcal{P}, \mathcal{A}$
\end{algorithmic}
\end{algorithm}

\section{Baseline Details}\label{app:baseline}
Unless otherwise specified, for all methods we use the LLM backbone and hyperparameters as described in the original papers.

RoG, G-Retriever, and GNN-RAG are each trained and evaluated separately on the 8:1:1 training split of each dataset (Family and FB15k-237) using a single NVIDIA H200 GPU, as described in Section~\ref{sec:data_overview}. For RoG, we use LLaMA2-Chat-7B as the LLM backbone, instruction-finetuned on the training split of Family or FB15K-237 for 3 epochs. The batch size is set to 4, the learning rate to $2 \times 10^{-5}$, and a cosine learning rate scheduler with a warmup ratio of 0.03 is adopted~\citep{linhao2024rog}. For G-Retriever, the GNN backbone is a Graph Transformer (4 layers, 4 attention heads per layer, hidden size 1024) with LLaMA2-7B as the LLM. Retrieval hyperparameters and optimization follow~\citet{he2024g}. For GNN-RAG~\citep{mavromatis2024gnn}, we use the recommended ReaRev backbone and sBERT encoder; the GNN component is trained for 200 epochs with 80 epochs of warmup and a patience of 5 for early stopping. All random seeds are fixed for reproducibility. For  PoG~\citep{chen2024pog}, StructGPT~\citep{jiang2023structgpt}, and ToG~\citep{Jiashuo2024ToG}, we use GPT-4o-mini as the underlying LLM, and the original prompt and generation settings from each method. For our proposed GR-Agent, we also adopt GPT-4o-mini as the backbone LLM to ensure fairness of comparison.

\section{Detailed Analysis of Other Rule Types}
\label{app:other_rules}

The \emph{Other} category in Table~\ref{tab:rule-stats} encompasses a broad range of logical rules that do not fall into standard symmetry, inversion, hierarchy, or composition classes. Below we summarize the main patterns observed, provide representative examples, and discuss their impact on model performance.

\paragraph{Longer Compositional Chains.} 
Rules involving three, 
\begin{align*}
    &r_1(x, y) \wedge r_2(y, z) \wedge r_3(z, w) \Rightarrow r_4(x, w)
\end{align*}

\paragraph{Triangle Patterns.}
Rules connecting three entities in a triangle motif, 
\begin{align*}
    &r_1(x, y) \wedge r_2(x, z) \Rightarrow r_3(y, z)
\end{align*}

\paragraph{Intersection Rules.}
Rules where multiple body atoms share the same argument, 
\begin{align*}
    &r_1(x, y) \wedge r_2(x, y) \Rightarrow r_3(x, y)
\end{align*}

\paragraph{Other  Patterns.}
Some rules do not exhibit simple interpretable motifs, involving unusual variable binding or rare predicate combinations. Like recursive rules (check AMIE3~\cite{lajus2020fast} for more details)





\newpage
\section*{NeurIPS Paper Checklist}

\begin{enumerate}

\item {\bf Claims}
    \item[] Question: Do the main claims made in the abstract and introduction accurately reflect the paper's contributions and scope?
    \item[] Answer:  \answerYes{}.
    \item[] Justification: The abstract and introduction clearly state the benchmark-construction methodology under KG incompleteness, the evaluation of existing methods, and the introduction of the GR-Agent framework.

\item {\bf Limitations}
    \item[] Question: Does the paper discuss the limitations of the work performed by the authors?
    \item[] Answer: \answerYes{}.
    \item[] Justification: The paper acknowledges limitations in the Conclusion, including evaluation scope (two KGs), robustness under incompleteness, and future extensions such as RL-based optimization.

\item {\bf Theory assumptions and proofs}
    \item[] Question: For each theoretical result, does the paper provide the full set of assumptions and a complete (and correct) proof?
    \item[] Answer: \answerNA{}
    \item[] Justification: The paper is empirical and methodological; it formalizes the agent decision process but does not introduce theorems requiring formal proofs.

    \item {\bf Experimental result reproducibility}
    \item[] Question: Does the paper fully disclose all the information needed to reproduce the main experimental results of the paper to the extent that it affects the main claims and/or conclusions of the paper (regardless of whether the code and data are provided or not)?
    \item[] Answer: \answerYes{} 
    \item[] Justification:  We describe dataset construction, splits, metrics, baseline configurations, and prompt settings, and provide an anonymized repository to enable reproducibility.

\item {\bf Open access to data and code}
    \item[] Question: Does the paper provide open access to the data and code, with sufficient instructions to faithfully reproduce the main experimental results, as described in supplemental material?
    \item[] Answer: \answerYes{} 
    \item[] Justification:     We release anonymized code and processed datasets with detailed instructions for reproduction in the supplementary material. This includes data preparation, hyperparameters, and commands to reproduce all reported results.


\item {\bf Experimental setting/details}
    \item[] Question: Does the paper specify all the training and test details (e.g., data splits, hyperparameters, how they were chosen, type of optimizer, etc.) necessary to understand the results?
    \item[] Answer: \answerYes{} 
        \item[] Justification: The paper specifies dataset splits, hyperparameters, optimizer, learning rate schedules, and early stopping criteria.


\item {\bf Experiment statistical significance}
    \item[] Question: Does the paper report error bars suitably and correctly defined or other appropriate information about the statistical significance of the experiments?
        \item[] Answer: \answerYes{} 
        \item[] Justification: We report results over multiple random seeds and include standard deviations as error bars in tables and figures, capturing variability due to initialization and train/validation splits.

\item {\bf Experiments compute resources}
    \item[] Question: For each experiment, does the paper provide sufficient information on the computer resources (type of compute workers, memory, time of execution) needed to reproduce the experiments?
    \item[] Answer: \answerYes{} 
    \item[] Justification: We provide the hardware (NVIDIA H200 GPUs, 80GB memory) and total compute time used for each experiment, along with estimated total compute across all runs, as detailed in Appendix~\ref{app:exp_details}.
    
\item {\bf Code of ethics}
    \item[] Question: Does the research conducted in the paper conform, in every respect, with the NeurIPS Code of Ethics \url{https://neurips.cc/public/EthicsGuidelines}?
    \item[] Answer: \answerYes{} 
    \item[] Justification: The research conforms to the NeurIPS Code of Ethics. All datasets are public benchmarks, no private or sensitive data is used, and anonymity is preserved for the review process.

\item {\bf Broader impacts}
    \item[] Question: Does the paper discuss both potential positive societal impacts and negative societal impacts of the work performed?
    \item[] Answer: \answerYes{} 
    \item[] Justification: We discuss both potential benefits (e.g., more robust KG-based reasoning systems for information retrieval and question answering).
    
\item {\bf Safeguards}
    \item[] Question: Does the paper describe safeguards that have been put in place for responsible release of data or models that have a high risk for misuse (e.g., pretrained language models, image generators, or scraped datasets)?
    \item[] Answer: \answerNA{} 
    \item[] Justification: The work does not involve releasing high-risk models or datasets such as pretrained LLMs or large-scale scraped corpora. All datasets are standard public benchmarks.

\item {\bf Licenses for existing assets}
    \item[] Question: Are the creators or original owners of assets (e.g., code, data, models), used in the paper, properly credited and are the license and terms of use explicitly mentioned and properly respected?
    \item[] Answer: \answerYes{} 
    \item[] Justification: All datasets (FB15k-237, Family) and baselines are properly cited.

\item {\bf New assets}
    \item[] Question: Are new assets introduced in the paper well documented and is the documentation provided alongside the assets?
    \item[] Answer: \answerYes{} 
    \item[] Justification: We introduce new benchmark variants. Documentation, data splits, and instructions for usage are provided with the released assets. Data is anonymized and follows licensing requirements of the original sources.

\item {\bf Crowdsourcing and research with human subjects}
    \item[] Question: For crowdsourcing experiments and research with human subjects, does the paper include the full text of instructions given to participants and screenshots, if applicable, as well as details about compensation (if any)? 
    \item[] Answer: \answerNA{} 
    \item[] Justification: The paper does not involve crowdsourcing or human-subject studies.

\item {\bf Institutional review board (IRB) approvals or equivalent for research with human subjects}
    \item[] Question: Does the paper describe potential risks incurred by study participants, whether such risks were disclosed to the subjects, and whether Institutional Review Board (IRB) approvals (or an equivalent approval/review based on the requirements of your country or institution) were obtained?
    \item[] Answer: \answerNA{} 
    \item[] Justification: The research does not involve human subjects and therefore does not require IRB approval.

\item {\bf Declaration of LLM usage}
    \item[] Question: Does the paper describe the usage of LLMs if it is an important, original, or non-standard component of the core methods in this research? Note that if the LLM is used only for writing, editing, or formatting purposes and does not impact the core methodology, scientific rigorousness, or originality of the research, declaration is not required.
    \item[] Answer: \answerYes{} 
    \item[] Justification: LLMs are used as part of the agent design for executing tool-based reasoning actions (exploration, grounding, synthesis). Their role in the methodology is fully described.

\end{enumerate}

\end{document}